# THE EFFECT OF SCENE CONTEXT ON WEAKLY SUPERVISED SEMANTIC SEGMENTATION

*Mohammad Kamalzare[1,2], Reza Kahani[2], Alireza Talebpour[1,2], and Ahmad Mahmoudi-Aznaveh[2]*

m.kamalzare@mail.sbu.ac.ir, r.kahani.sbu@gmail.com, talebpour@sbu.ac.ir, a_mahmoudi@sbu.ac.ir
[1]Department of Computer Science and Engineering, Shahid Beheshti University, Tehran, Iran
[2]Cyberspace Research Institute, Shahid Beheshti University, Tehran, Iran

**ABSTRACT**

Image semantic segmentation is parsing image into several partitions in such a way that each region of which involves a semantic concept. In a weakly supervised manner, since only image-level labels are available, discriminating objects from the background is challenging, and in some cases, much more difficult. More specifically, some objects which are commonly seen in one specific scene (e.g. "train" typically is seen on "railroad track") are much more likely to be confused. In this paper, we propose a method to add the target-specific scenes in order to overcome the aforementioned problem. Actually, we propose a scene recommender which suggests to add some specific scene contexts to the target dataset in order to train the model more accurately. It is notable that this idea could be a complementary part of the baselines of many other methods. The experiments validate the effectiveness of the proposed method for the objects for which the scene context is added.

## 1. INTRODUCTION

Semantic image segmentation is the task of pixel-level labeling of an image. Lately, fully supervised approaches have reached favorable achievements, however, these methods rely on huge pixel level annotated data which is exhausting and time-consuming [1]. This is the main motivation to explore the possibility of training the systems using weak supervision setting called weakly supervised semantic segmentation [2-5], in which each training image is annotated by image-level labels. It is obvious that in this setting, objects can easily be confused with the background in the training stage, especially the objects that typically occur in a specific scene (e.g. "boat"). In this paper, we propose the incorporation of scene information to tackle this problem. To this end, a co-occurrence matrix is computed between the target objects and a scene dataset, and some of the scene classes are recommended to be added to the target dataset lastly. The experiments demonstrate that adding appropriate scene classes can considerably help the weakly supervised network to learn object boundaries more precisely.

## 2. RELATED WORKS

In the past few years, image semantic segmentation has attracted a lot of attention. Most of the existing works model the task as a fully supervised problem. It is to be noted that the segmentation methods that are based on Fully Convolutional Networks (FCN) also promote the level of the task, which introduce an end-to-end manner. Numerous work have been proposed with promising results [6-10], nevertheless, these methods need a huge annotated data in the training phase which is time-consuming and labor-intensive.

Recently, weakly supervised methods begun to address image semantic segmentation problems in which each training image is only annotated by image-level labels [5, 11-13].

Simonyan *et al.* [14] extracted a saliency map by estimating the gradient of class scores from classification ConvNets (Convolutional Networks) regarding the input image. They refined the extracted map by using a graph cut method. Zhou *et al.* [11] improved [14] by introducing Class Activation Mapping (CAM) technique which was acquired from global average pooling layer (GAP) of an classification ConvNet. Although these models, propose worthy localization cues, they may miss many parts of the objects. It is because the methods rely on discriminative features that separate each object from the others in image classifiers.

Kolesnikov *et al.* [15] introduced a new loss function which was composed of three components: Seed, Expand, and Constrain (SEC) to train an FCN in a weakly supervised way. They exploited CAM to produce a weak localization cue of objects and proposed a global weighted rank pooling to measure consistency between the image-level label and the corresponding mask. Moreover, they employed Conditional Random Field (CRF) to improve the discontinuity of the segments in the masks. In [13], a method is proposed to incorporate foreground/background knowledge extracted from the convolutional layers of a network, and to fuse them with the information attained from a weakly supervised localization network. Using weak supervision in these methods causes them to be incapable of distinguishing the object from the background in some cases. For example, the methods usually label "sea" as "boat". Some of these failure cases occur mainly because the object is mostly seen in just one scene. More specifically, some objects like "boat" are mostly observed in a specific background like "sea", this causes the model to consider the background as an object. Solving this problem is our main motivation.

In this paper, we propose a method to resolve the problem mentioned previously. Our method recommends some scene contexts to be added to the training data. The scene context information prevents the model from confusing the background and the object in the cases are included in the problem.

Shen *et al.* [5] developed a pipeline in which they utilized SEC method to generate the masks, and to finally train an FCN in a fully supervised manner. Their method heavily relies on the quality of masks which are generated by using the SEC, thus, the accuracy of the final model is directly affected by that of the SEC. As a result, it can be concluded that more precise masks generated by the SEC model can result in a more accurate final model.

## 3. PROPOSED METHOD

One of the main challenges in weakly supervised semantic segmentation methods is their limited supervision. In other words, in these methods, the model in the training phase might be misled to confuse classes. This is common to increase the amount of training data to resolve this problem [5, 15]. However, extending the amount of data may be insufficient in some specific classes due to the relationship between the class and its scene. A specific class may happen almost in one scene (e.g. "boat" is often appeared in "sea"). Therefore, the model in the training phase could not distinguish between the class and its background. Consequently, this causes the scene to be learned as a part of the object class. (see Figure 1)

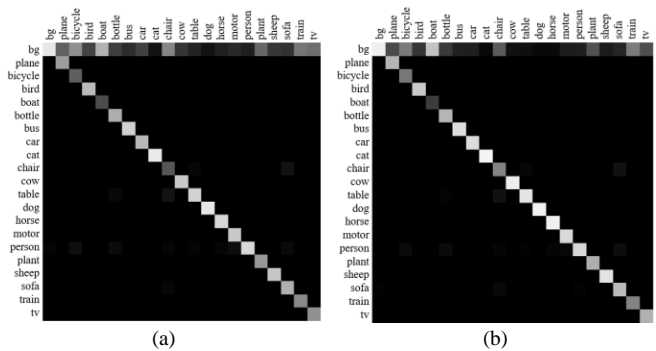

(a) (b)

**Figure 1. (a) is a pixel-wise confusion matrix of SEC [15] and (b) is that of [5] among Pascal VOC 2012 classes**

Figure 1 shows the confusion matrix over the training set of Pascal VOC 2012 of SEC [15] and [5] methods. For each matrix, each cell is normalized as follows:

$$c_{ij} = v_{ij} \Big/ \sum_{h=1}^{N} v_{ih} \qquad (1)$$

where $v_{ij}$ is the number of pixels belonging to *i*-th class that is predicted as *j*-th class, and *N* is the number of classes.

The matrix, which is inferred from the SEC method, demonstrates that in the case of "boat", almost 70% of the background pixels are predicted as "boat". It also shows that in the case of "train", more than 45% of the background pixels are predicted as "train". Although the presented method in [5] has extended the training data by adding samples to all the 21 classes of Pascal VOC, it has also a similar results in the case of "boat" and "train". It can be concluded that although increasing the number of samples for each class can be effective for some classes such as "bottle", some other classes like "boat" and "train" remained almost unchanged in terms of precision. As it is mentioned before, this problem happens because the target object mostly occurs in a same scene (see Figure 2).

To overcome this problem, we propose to add scene contexts information in the training phase with a method consisting of the following steps:

1. **Preparing image-level labeled dataset:** Gathering images for the target classes. Each image should be labeled by a list of objects presented in.
2. **Computing co-occurrence matrix *M*:** In order to find the scenes which often occurred with each object, we propose to compute co-occurrence matrix *M* between the scenes and the object classes. In the experiments, the Places2 dataset [16] is used as the scene dataset. Suppose that *M* be a $H \times N$ matrix. *H* is the number of scene classes, and *N* is the number of objects. The occurrence of *i*-th scenes presented in *j*-th object images is denoted by $M_{ij}$.
3. **Finding the candidate scene classes:** For this purpose, we consider a scene class ζ as a candidate for target object δ if: (1) the scene ζ rarely occurs in any other object images. (2) the object δ mostly occurs in the scene ζ. we defined scene scores *A* as follows:

$$A = \{\frac{m_{ij}}{\sum_{k=1}^{H} m_{kj}} : \frac{m_{ij}}{\sum_{z=1}^{N} m_{iz}} > T, 1 \le i \le H, 1 \le j \le N\} \quad (2)$$

Where *T* is a threshold that satisfies condition (1) in the candidate definition. The scene classes which their scores are among the top *n* scores of set *A* are considered as the candidates. Each of the candidate scenes indicates which scene classes are more suitable to be added to the target object classes. It also implies that which class will be more affected.

4. **Data cleaning:** Let $Q_i$ be the set of all the images of the candidate scenes that are proposed to be added to *i*-th target object. We define $G_i \subseteq Q_i$ as a clean set of scene images for *i*-th target object that can be achieved by removing all of the images which contain the target objects from $Q_i$, therefore, the images of $G_i$ are added to the target dataset as the scene for the *i*-th target object (denoted by $S_i$). It should be noted that adding $S_i$ could be effective when $\|G_i\|$ is not very small.
5. **Training a weakly supervised model:** An object localizer network is trained after adding the selected scene classes to the target dataset. For this purpose, the SEC network is utilized [15].

In the experiments, it is shown that the accuracy of the proposed method was impressively improved for the objects that the scene class is added (i.e. "boat" and "train"). We also show that our method could successfully improve the total accuracy of the SEC [15]. It should be mentioned that the method is applicable for any weakly supervised semantic segmentation method.

## 4. EXPERIMENTAL RESULTS

In this section, first, the employed dataset is introduced, then, our experiment setup is explained, and finally, our results and comparisons with other methods are presented.

**Dataset:** We used the Pascal VOC 2012 image segmentation benchmark [17] to evaluate our method. The dataset contains 21 semantic classes including the background and it is split into three sections: training set (1464 images), validation set (1449 images) and testing set (1456 images). The standard Pascal VOC metric was used as the evaluation measure which is mean intersection over union (mIoU) [17].

We also utilized the train set of Places365-Standard, a subset of Places2 which has ~1.8 million images from 365 scene categories and there are at most 5000 images per category, to extract scene images. The Places2 dataset is constructed to resolve the problems are dependent on scene data. This dataset is a repository of about 10 million images of scene semantic categories, and the performance of CNNs are trained on this dataset demonstrates that this repository significantly outperforms the performance of tasks such as visual object and scene recognition. In this work the pre-trained model "Places365-VGG", which is offered by [16] and has the highest performance in scene classification task, is supplied to identify the corresponding background of each PASCAL dataset object. To elucidate more, we feed each picture of PASCAL dataset to the Places365-VGG model and store the score of each scene class. Actually, the set of these scores ultimately shows that each object often occurs in which scene or scenes, and is exploited to compute co-occurrence matrix M.

**Experiment Setup:** We employed the SEC method by using resnet-50 [18] as the base network as same as [5]. In order to clarify the impact of the added scene classes more precisely, we also reported the SEC performance using resnet-50 base network which is trained on the web data in the tables denoted by "SEC-web".

To compute the co-occurrence matrix *M*, and to obtain the scene classes, the pre-trained model "Places365-VGG" [16] was used. More specifically, the top 5 predictions of the network were taken for each image of Pascal VOC 2012 (training set) and the matrix *M* was computed. Then, we selected the images of *n*=11 top scene classes from scene dataset Places2 [16] and cleaned them. Consequently, the scene classes were considered, for which $\|G_i\|$ was not too small, as selected. We picked at last 5000 images from each clean set (i.e. clean set *G* for target objects: "boat" and "train") and add them into target dataset. The total extracted images from the scene dataset finally reached to 10000.

To train the SEC model the filtered web data of [5] was used (we similarly train an initial SEC on the training set of Pascal VOC 2012 to filter the original web data), plus the scene data, which totally contains 81049 images of 23 classes (including Pascal VOC + the scenes). In Eq. (2), we set $T = 0.3$.

**Experiments and Results:**

Table 1 and

Table 2 display the results on Pascal VOC 2012 validation and test set in comparison with the SEC [15] and "SEC-web", respectively. Accordingly, the results of the two Pascal VOC classes for which we added scene data, are significantly outperformed. Additionally, our method improved the baseline (SEC-web) in terms of mean IoU.

The qualitative results of our method are illustrated in Figure 2. The first four rows show our results for some of the failure cases of the other works on Pascal VOC validation set. The last two rows are some random samples for "railroad" and "sea" subjects downloaded from Flickr website. The results show that our method not only distinguishes the objects form their common backgrounds, but also its false alarms, that is the regions not containing the target objects, are very smaller than the other methods. The suppression of false alarms becomes more important in the practical applications, where the data is imbalanced. As it is shown in the Table 1, the improvement regarding to the baseline (SEC-web) for classes "boat" and "train" are up to 26% and 20%, respectively.

It can be concluded that adding the scene context helps the model to conquer certain parts of the shortcomings that arose from image-level supervision of weakly supervised segmentations. We think that this method can be appended to the other weakly supervised systems to cover the missing supervision.

**Table 1. Results on the Pascal VOC 2012 validation set**

| Method | bk | plane | bicycle | bird | boat | bottle | bus | car | cat | chair | cow | table | dog | horse | motor | person | plant | sheep | sofa | train | tv | mean |
|---|---|---|---|---|---|---|---|---|---|---|---|---|---|---|---|---|---|---|---|---|---|---|
| Pinheiro *et al.* [19] | 77.2 | 37.3 | 18.4 | 25.4 | 28.2 | 31.9 | 41.6 | 48.1 | 50.7 | 12.7 | 45.7 | 14.6 | 50.9 | 44.1 | 39.2 | 37.9 | 28.3 | 44.0 | 19.6 | 37.6 | 35.0 | 36.6 |
| SEC [15] | 82.4 | 62.9 | 26.4 | 61.6 | 27.6 | 38.1 | 66.6 | 62.7 | 75.2 | 22.1 | 53.5 | 28.3 | 65.8 | 57.8 | 62.3 | 52.5 | 32.5 | 62.6 | 32.1 | 45.4 | 45.3 | 50.7 |
| Saleh *et al.* [13] | 82.2 | 59.5 | **27.4** | 66.7 | 25.2 | 44.1 | 71.1 | 55.1 | 71.9 | 19.7 | 52.3 | 36.7 | 65.6 | 59.4 | 62.8 | 55.3 | 32.3 | 65.5 | **34.3** | 43.4 | 38.8 | 50.9 |
| SEC-web | 81.3 | 51.9 | 25.1 | 59.2 | 20.5 | 54.0 | 73.6 | 67.0 | 73.7 | 18.8 | 58.4 | 30.8 | 65.4 | 53.2 | **63.2** | 59.2 | 33.3 | 59.3 | 32.9 | 37.4 | 53.2 | 51.0 |
| SEC-web+crf | 83.7 | **66.5** | 24.0 | 75.6 | 21.7 | **56.0** | 74.4 | 66.4 | **77.7** | 19.9 | **62.9** | 30.0 | 69.3 | **62.0** | 62.0 | **61.7** | 33.4 | 68.5 | 33.0 | 38.8 | **56.1** | 54.4 |
| Ours | 84.4 | 51.9 | 23.9 | 57.8 | **47.2** | 54.7 | 72.4 | 66.3 | 73.0 | 19.0 | 56.2 | 30.5 | 66.9 | 51.7 | 61.6 | 57.7 | 35.6 | 60.5 | 32.3 | **57.6** | 51.2 | 53.0 |
| Ours + crf | **86.7** | **66.5** | 22.8 | **76.8** | 43.7 | **56.0** | **75.1** | **66.7** | **78.0** | 19.1 | 61.2 | 29.2 | **70.6** | 59.1 | 61.6 | 60.0 | **38.1** | **71.5** | 31.9 | 51.2 | 54.3 | **56.2** |

**Table 2. Results on the Pascal VOC 2012 test set**

| Method | bk | plane | bicycle | bird | boat | bottle | bus | car | cat | chair | cow | table | dog | horse | motor | person | plant | sheep | sofa | train | tv | mean |
|---|---|---|---|---|---|---|---|---|---|---|---|---|---|---|---|---|---|---|---|---|---|---|
| Pinheiro *et al.* [19] | 74.7 | 38.8 | 19.8 | 27.5 | 21.7 | 32.8 | 40.0 | 50.1 | 47.1 | 7.2 | 44.8 | 15.8 | 49.4 | 47.3 | 36.6 | 36.4 | 24.3 | 44.5 | 21.0 | 31.5 | 41.3 | 35.8 |
| Shimoda *et al.* [20] | 78.1 | 43.8 | 26.3 | 49.8 | 19.5 | 40.3 | 61.6 | 53.9 | 52.7 | 13.7 | 47.3 | 34.8 | 50.3 | 48.9 | 69.0 | 49.7 | 38.4 | 57.1 | 34.0 | 38.0 | 40.0 | 45.1 |
| SEC [15] | 83.5 | 56.4 | 28.5 | 64.1 | 23.6 | 46.5 | **70.6** | 58.5 | 71.3 | **23.2** | 54.0 | 28.0 | 68.1 | 62.1 | 70.0 | 55.0 | 38.4 | 58.0 | 39.9 | 38.4 | 48.3 | 51.7 |
| Saleh *et al.* [13] | 83.5 | 60.8 | **29.8** | 66.6 | 23.2 | 52.1 | 69.3 | 53.8 | 70.4 | 19.1 | 56.8 | 40.1 | **71.0** | 59.7 | **71.4** | 54.9 | 33.9 | **71.2** | **40.5** | 35.4 | 41.9 | 52.6 |
| Ours | 84.9 | 50.3 | 26.9 | 57.0 | **41.3** | 60.6 | 68.8 | **66.9** | 69.6 | 20.4 | 58.8 | 41.8 | 64.4 | 54.5 | 70.8 | 59.9 | 38.1 | 57.9 | 36.3 | **53.2** | 50.3 | 53.9 |
| Ours + crf | **87.3** | **65.8** | 27.9 | **74.3** | 38.5 | **63.7** | 69.9 | 65.3 | **72.2** | 21.2 | **64.8** | 42.1 | 67.2 | **67.1** | 71.0 | **62.0** | 40.2 | 69.1 | 35.3 | 43.1 | **54.1** | **57.2** |

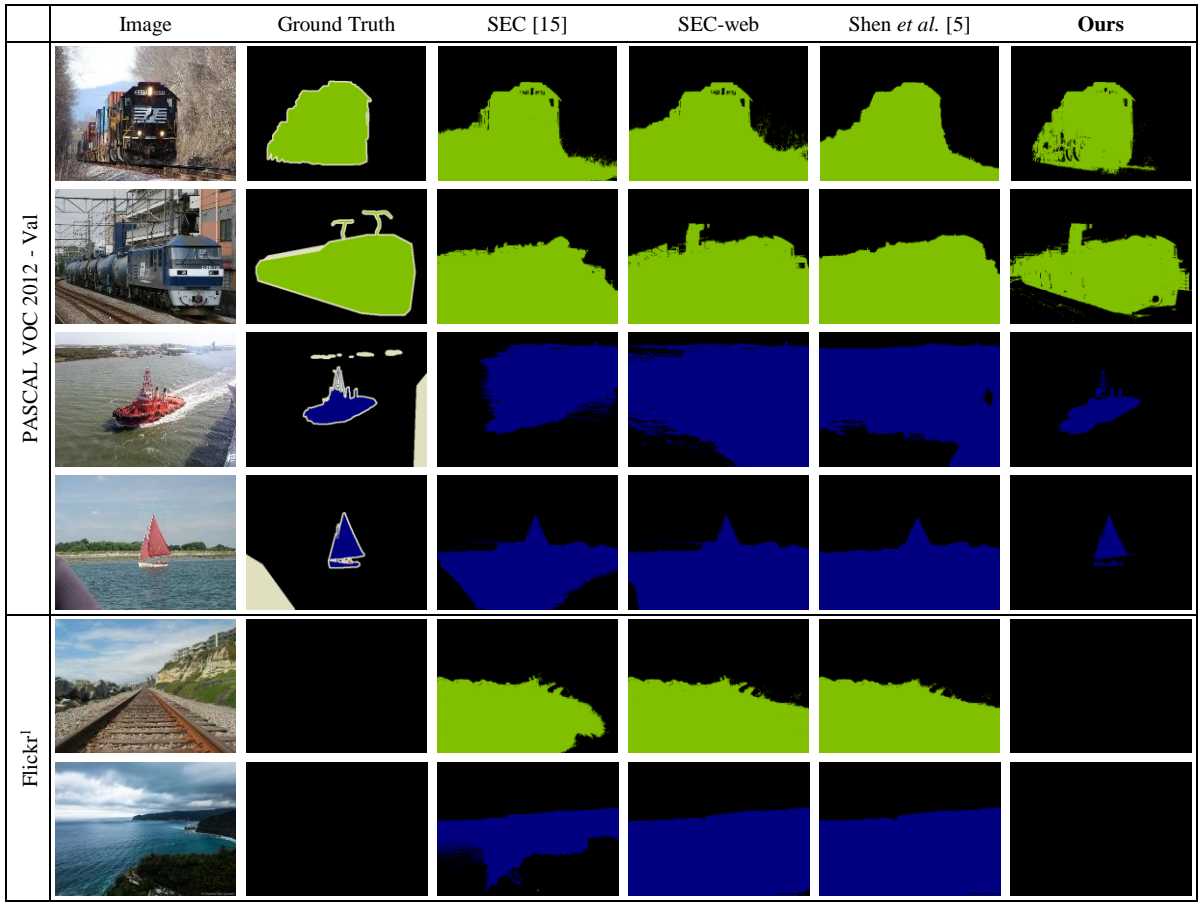

**Figure 2.** Qualitative results on Pascal VOC 2012 validation set for "boat" and "train" classes. (best view in color)

**CONCLUSION**

In this paper, a method was proposed to overcome certain parts of the shortcomings that arose from image-level supervision of weakly supervised segmentations by adding the scene context information. Accordingly, we proposed a system which recommends the appropriate scene context to be added to the target dataset. The problem we solved was the confusion of objects and the background, which is intensified for the objects that mostly occur in one specific scene.

Furthermore, the experiments demonstrated the effectiveness of our method. The improvement achieved regarding the baseline (SEC-web) for classes "boat" and "train" were up to 26% and 20%, respectively.

---

[1] www.flickr.com